\documentclass[runningheads]{llncs}

\usepackage{eccv}

\usepackage{eccvabbrv}
\usepackage{comment}
\usepackage{graphicx}
\usepackage{booktabs}
\usepackage{algorithmic}
\usepackage{algorithm}
\usepackage{multirow}
\usepackage{svg}
\svgsetup{inkscapelatex=false}
\usepackage[accsupp]{axessibility}  

\usepackage{hyperref}

\usepackage{orcidlink}

\begin{document}

\title{Exploring 3D Face Reconstruction and Fusion Methods for Face Verification: A Case-Study in Video Surveillance}




\titlerunning{3DFR and Fusion Methods for Face Verification in Video Surveillance}

\author{Simone Maurizio La Cava\inst{1}\orcidlink{0000-0002-6344-1845} \and
Sara Concas\inst{1}\orcidlink{0000-0001-8114-0686} \and
Ruben Tolosana\inst{2}\orcidlink{0000-0002-9393-3066} \and
\\
Roberto Casula\inst{1}\orcidlink{0000-0003-3810-5935} \and
Giulia Orrù\inst{1}\orcidlink{0000-0002-7802-2483} \and
Martin Drahansky\inst{3}\orcidlink{0000-0002-9321-7385} \and \\
Julian Fierrez\inst{2}\orcidlink{0000-0002-6343-5656} \and 
Gian Luca Marcialis\inst{1}\orcidlink{0000-0002-8719-9643}
}

\authorrunning{S.~M.~La Cava et al.}
\institute{University of Cagliari, Cagliari, Italy 
\email{\{simonem.lac,sara.concas90c,roberto.casula,giulia.orru,marcialis\}@unica.it} \and
Autonomous University of Madrid, Madrid, Spain \\
\email{\{ruben.tolosana,julian.fierrez\}@uam.es} \and
Masaryk University, Brno, Czech Republic \\
\email{drahansky@sci.muni.cz}}

\maketitle

\begin{abstract}

3D face reconstruction (3DFR) algorithms are based on specific assumptions tailored to distinct application scenarios. These assumptions limit their use when acquisition conditions, such as the subject's distance from the camera or the camera's characteristics, are different than expected, as typically happens in video surveillance. Additionally, 3DFR algorithms follow various strategies to address the reconstruction of a 3D shape from 2D data, such as statistical model fitting, photometric stereo, or deep learning. In the present study, we explore the application of three 3DFR algorithms representative of the SOTA, employing each one as the template set generator for a face verification system. The scores provided by each system are combined by score-level fusion. 
We show that the complementarity induced by different 3DFR algorithms improves performance when tests are conducted at never-seen-before distances from the camera and camera characteristics (cross-distance and cross-camera settings), thus encouraging further investigations on multiple 3DFR-based approaches.

\end{abstract}

\section{Introduction}
\label{sec:intro}
In the last few years, much attention has been paid to the generation of face synthetic images \cite{melzi2024frcsyn,deandres2024frcsyn,shahreza2024sdfr}, in particular, 3D data \cite{richardson20163d,sun2022controllable}. The acquisition of 3D data has been proven to be robust to adverse factors of uncontrolled environments, such as unfavorable illumination conditions and non-frontal poses of the face \cite{2D3Dsurvey,robust,robustsurvey}.  
Moreover, high accuracy and efficiency can be achieved when comparing faces due to the complementary information of shape and texture \cite{3Dstereo,ICPrecognition}. However, in comparison to standard 2D images, the acquisition of such 3D data requires a much more complex enrolment process and expensive hardware \cite{la20233d,uncalibrated}. 
This is the main reason why current face recognition technology is still mostly based on the acquisition of 2D face images, considering popular and cost-effective 2D acquisition devices. 

\begin{figure}[h]
\centering
\includegraphics[width=0.6\linewidth]{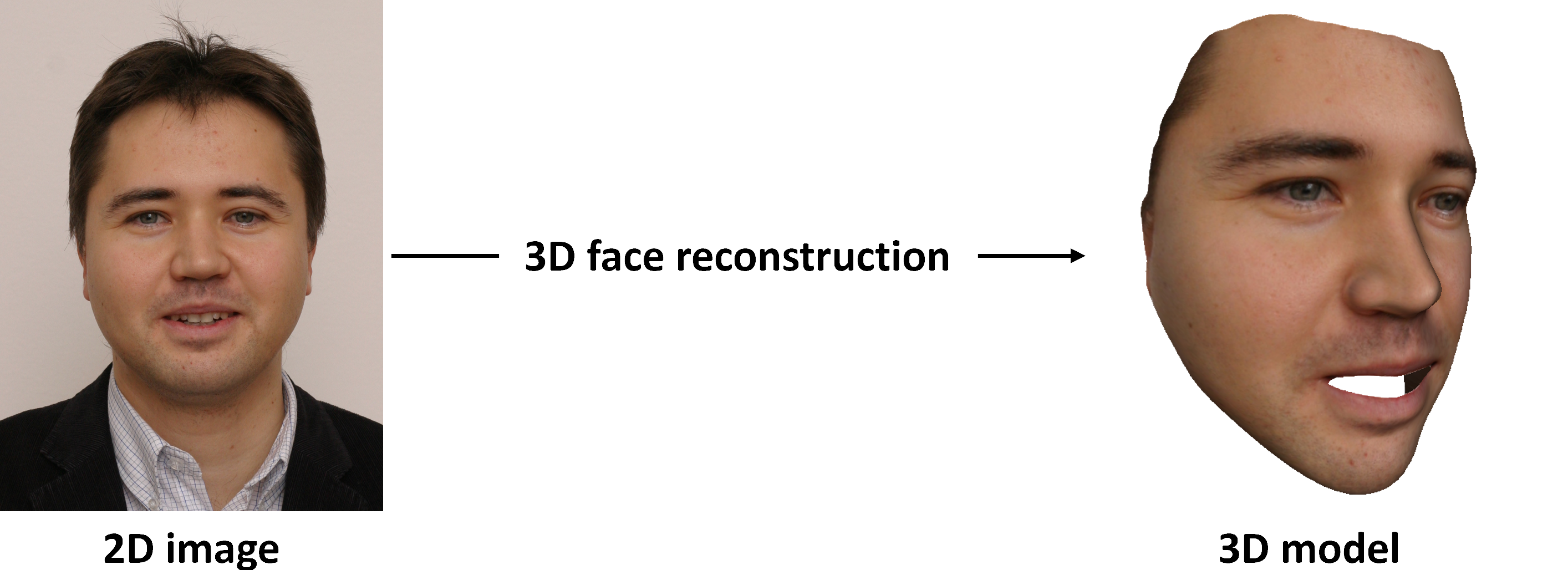}
	\caption{Example of 3D face reconstruction from a single 2D image using EOS \cite{eos}.}
	\label{fig:3DreconstructionExample}
\end{figure}

Approaches based on 3D face reconstruction (3DFR) from 2D images and videos can be a good solution to overcome the limits of 2D data, combining the ease of acquiring 2D data with the robustness of 3D facial models \cite{la20223d}. An example of this can be seen in Figure \ref{fig:3DreconstructionExample}.
3DFR algorithms have proven to be particularly suitable in video surveillance scenarios, where a face acquired in unconstrained and different acquisition settings, even with arbitrary pose and distance, is compared to the identity using a reference image, \textit{\eg}, mugshot or identity document \cite{la20233d}.

However, it is important to highlight that each 3DFR algorithm is usually designed for a specific scenario, such as recognition under certain environmental and technological conditions, or even for applications not strictly related to facial recognition \cite{la20233d,uncalibrated}. Therefore, they are not required to generalize well in other contexts. Accordingly, the designer "maximizes" specific characteristics, such as fine or grained details or even single facial components, useful for the specific application scenario. With this in mind, in the present study, we intend to answer the following research question: is it possible to improve face recognition performance in video surveillance scenarios through the fusion of the complementary information provided by different 3DFR algorithms? 

In order to answer this research question, we propose to jointly use different 3DFR algorithms for individually aiding the training of multiple classifiers, finally adopting a score-level fusion rule to moderate their matching score \cite{concas2022experimental} before making the final decision in a verification task, \textit{\ie}, determine whether the subject identity associated with the reference data is the same represented in the probe data based on the resulting \textit{a posteriori} probability. 
To summarize, the main contributions of the present study are the following: \textit{(i)} the analysis and exploitation of the complementary information provided by different 3DFR algorithms (EOS, 3DDFA v2, and NextFace), not strictly proposed for recognition purposes, to enhance a video surveillance scenario using two distinct deep-learning systems with significantly different architectural complexities; \textit{(ii)} the exploration of the effectiveness, advantages, and disadvantages of various score-level fusion methods to take advantage of the complementary information provided by different 3DFR algorithms; and \textit{(iii)} the assessment of the robustness of the proposed approach in challenging conditions, \textit{\ie}, when dealing with data acquired in a different setting, concerning acquisition distance and surveillance camera, from the one considered for the system design.

The rest of the paper is organized as follows. Section \ref{sec:relatedwork} discusses state-of-the-art 3D face reconstruction in video surveillance and fusion methods in face recognition. Section \ref{sec:proposed} describes our proposed approach based on 3DFR algorithms and fusion methods. Section \ref{sec:expsettings} reports the experimental protocol considered to investigate the effectiveness in a surveillance scenario. The experimental results are reported in Section \ref{sec:results}. Finally, conclusions are drawn in Section \ref{sec:conclusion}.

\section{Related Works} \label{sec:relatedwork}

\subsection{Exploiting 3DFR for Face Recognition}
\label{subsec:personalRecognition}

3DFR has been mainly proposed to increase the robustness of face recognition to faces acquired with various view angles, as this is the typical acquisition observed in unconstrained scenarios, with non-frontal and looking-down probe faces due to the non-cooperation of the subjects \cite{HU201746,la20223d}.
However, the performance improvement between 2D and 3DFR strongly depends on the approaches chosen for its application to the face recognition task. These are usually divided into two main categories, namely, model- and view-based approaches \cite{la20223d}. 

The first one synthesizes frontal faces from the 2D images containing non-frontal views. The normalized (or "frontalized") faces are then compared to the frontal ones to determine the subjects' identity \cite{HanJain2012}.
This category is prone to produce textural artifacts in the synthesized frontal images \cite{hassner2015effective,Cao2020,RotateAndRender}; thus, it is usually considered in the so-called face identification task rather than for highly accurate authentication \cite{HanJain2012}.

The second one adapts the 2D images containing frontal faces to non-frontal ones; in other words, lateral views derive from the 3D template (or model) \cite{HanJain2012}. Additionally, the 3D facial model can be projected to various poses in the 2D domain to enhance the representation capability of each subject, considering them as synthesized templates \cite{Zhang2008,Liang2018}. 
In general, view-based approaches are more expensive than model-based ones in terms of computational and storage costs. Thus, they are usually used in the verification task when high reliability is required \cite{HanJain2012}.

\subsection{3DFR in Video Surveillance Scenarios}
\label{subsec:3dfrANDvsurveillance}

Video surveillance scenarios are characterized by faces captured at an extensive range of lighting, pose, and scale due to environmental conditions and the subject's cooperation level. In other words, we must rely on uncalibrated images.

The 3D reconstruction from uncalibrated images is an inherently ill-posed problem: the facial geometry, the pose of the head, and its texture must be recovered from a single picture, leading to an undetermined problem, while different 3D faces could generate the same 2D image. 
Therefore, the research community has proposed that prior knowledge should be included to estimate the actual 3D geometry \cite{uncalibrated}. This information can be added through three main ways: photometric stereo, statistical model fitting, and deep learning. 

Through the first approach, a 3D template is combined with photometric stereo methods to estimate the facial surface model. Despite the capability of acquiring fine details, the reconstruction is typically performed through multiple images, thus further constraining the problem \cite{3Dstereo,uncalibrated}. 

Statistical model fitting consists of adapting a 3D facial model built from a set of 3D facial scans to the input images. In this context, the most commonly used statistical model is the 3DMM (3D Morphable Model), which consists of a shape model and, optionally, an albedo model, separately constructed using Principal Component Analysis (PCA) \cite{10.1145/3596711.3596730}. Despite providing generally convincing results and lower computational complexity, this approach focuses primarily on global characteristics rather than fine details \cite{uncalibrated}.

Similarly, the deep learning approach maps 2D to 3D information through a trained deep neural network. This approach provides more detailed results than the previous one at the cost of requiring more 3D scans to train the network \cite{uncalibrated}. 

Each of the approaches presented above performs 3DFR from uncalibrated images, as those typically acquired in video surveillance scenarios, following assumptions to make an intrinsically ill-posed problem into a manageable one \cite{geng2020towards,uncalibrated}. These assumptions and related pros and cons point out an intrinsic complementarity. Moreover, some studies highlight that it is possible to reconstruct highly detailed 3D faces even with a single image by combining the prior knowledge of the global facial shape encoded in the 3DMM and refining it through a photometric or a deep learning approach \cite{la20233d}. This further supports the hypothesis of the complementarity of the 3DFR methods. 
Therefore, combining face recognition systems based on multiple 3DFR algorithms could increase the generalization capabilities of the systems. To achieve this, we propose fusion techniques to exploit the complementary information provided by different 3DFR algorithms.

\subsection{Ensemble and Fusion Methods for Face Recognition} \label{sec:fusion}
Ensemble methods and multi-modal or uni-modal fusion approaches are constantly being exploited in many fields of pattern recognition to improve the ability to generalize and deal with intra-class variations and inter-class similarity \cite{vishi_evaluation_2018}. In biometrics, fusion can be performed at various levels, including sensor, feature, score, and decision levels \cite{Ross2009,Noore2009,Osadciw2009,peng2014multimodal}.

Among others, score-level fusion allows the exploitation of several sources of information without increasing the system's complexity (as in the case of sensor-level and feature-level fusion) and without relying only on the binary outcome, as in the case of decision-level fusion. A comprehensive set of previous studies supports this since the earlier attempts at face recognition \cite{marcialis2004fusion,dass2005principled,Ross2009,punyani_evaluation_2017}. 

\section{Proposed Method}
\label{sec:proposed}

\begin{figure*}[!h]
\centering
\includegraphics[width=\textwidth]{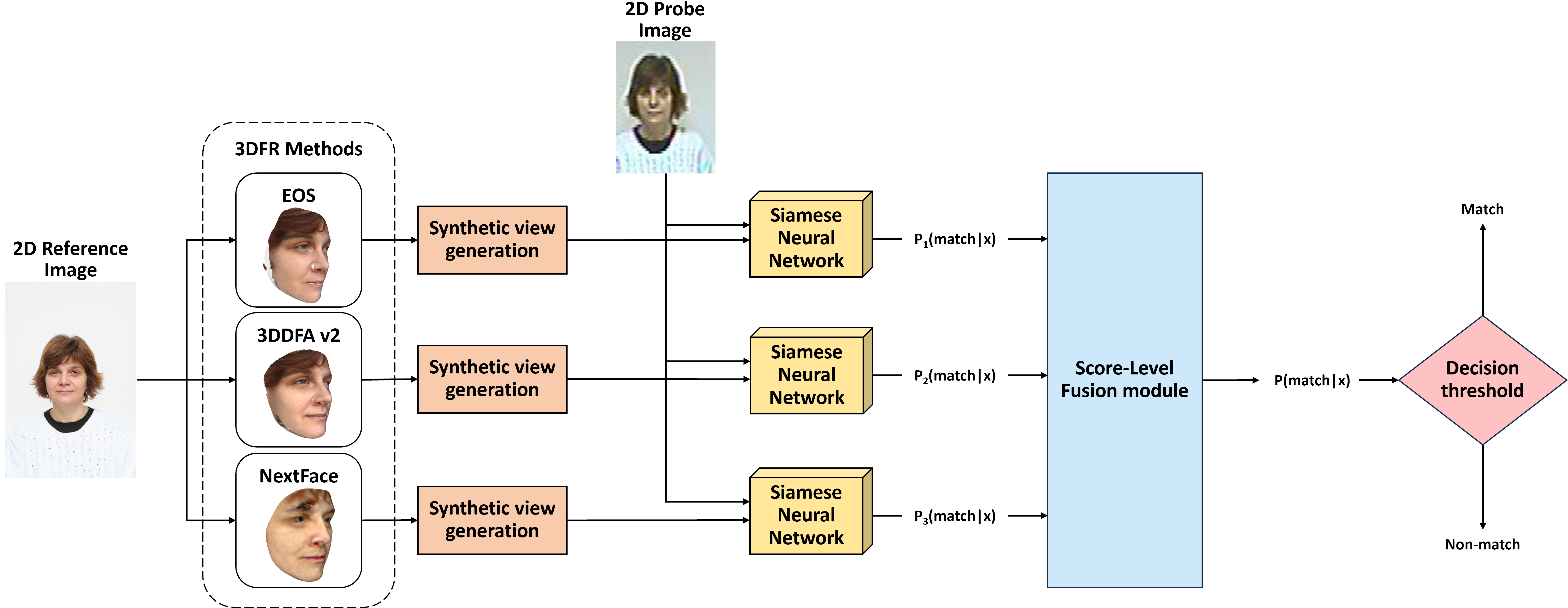}
\caption{Proposed method. The synthetic view generation produces a 2D image from the 3D template (\textit{\ie}, with various view angles obtained from gallery enlargement during the system's training, only in frontal view during inference). The Siamese Neural Networks (same architecture) provide complementary information as they are enhanced through different 3DFR algorithms (EOS, 3DDFA v2, or NextFace). The example images are from the SCface database \cite{SCFace}.}\label{fig:proposedsystem}
\end{figure*}

Figure \ref{fig:proposedsystem} provides a graphical representation of the proposed method to improve the performance of face verification in video surveillance scenarios, \textit{\ie}, determine whether the identity in a surveillance image (\textit{\ie}, probe) matches that represented in the reference data (\textit{\ie}, mugshot or template). In the following, we summarize the main modules.

\begin{figure}[t]
	\centering
 
	\includegraphics[width=0.6\linewidth]{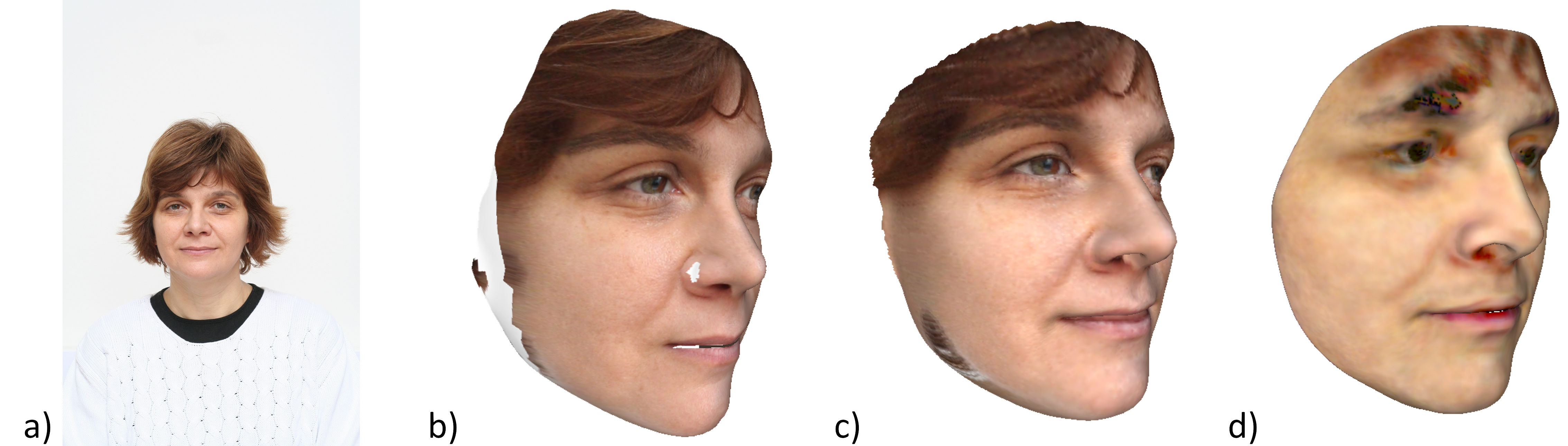}
	\caption{Examples of personalized 3D templates generated from a mugshot in the SCface database \cite{SCFace} (a), through EOS \cite{eos} (b), 3DDFA v2 \cite{guo2020towards} (c), and NextFace \cite{dib2021practical} (d).}
	\label{fig:3Dmodels}
\end{figure}

\textbf{3DFR methods:} We have chosen three types of state-of-the-art 3DFR algorithms, each representative of one or a combination of the previously described approaches for reconstructing the 3D templates from high-quality reference data (\textit{\ie}, frontal mugshot images):
\begin{itemize}
 \item EOS \cite{eos}: based on 3DMM and proposed for reconstructing 3D faces from videos and images in time-critical applications
 (Figure \ref{fig:3Dmodels}, b)\footnote{Original implementation from \url{https://github.com/patrikhuber/eos}}.
 \item 3DDFA v2 \cite{guo2020towards}: based on a lightweight network for regressing the 3DMM parameters and proposed for 3D dense face alignment (Figure \ref{fig:3Dmodels}, c)\footnote{Original implementation from \url{https://github.com/cleardusk/3DDFA\_V2}}.
 \item NextFace \cite{dib2021practical}: based on the combination of a statistical 3DMM and a photometric approach for making 3D reconstruction robust to light conditions (Figure \ref{fig:3Dmodels}, d)\footnote{Original implementation from \url{https://github.com/abdallahdib/NextFace}}.
\end{itemize}

\begin{figure}[t]
	\centering
	\includegraphics[width=0.9\linewidth]{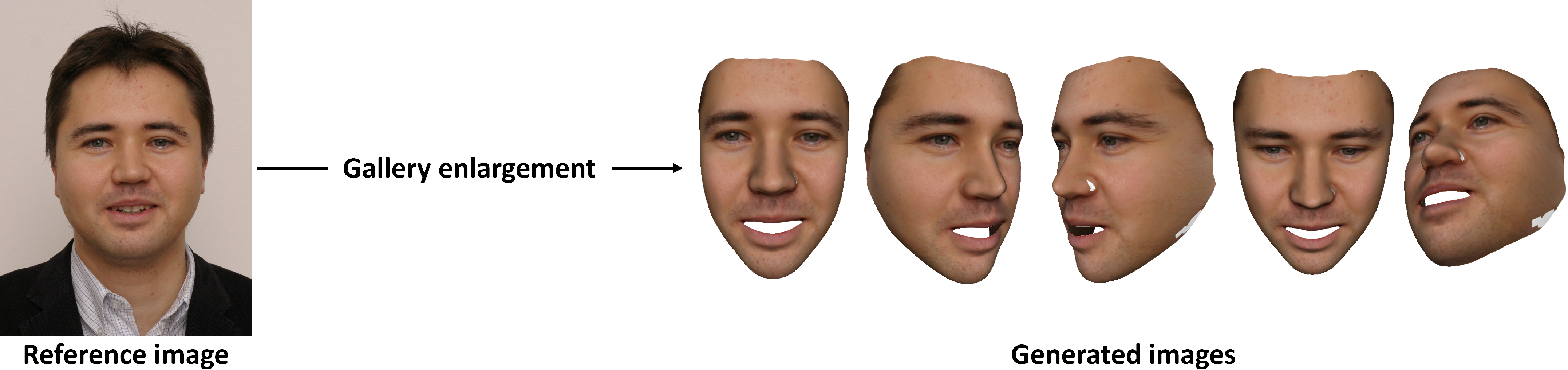}
	\caption{Example of gallery enlargement from personalized 3D template obtained from a mugshot in the SCface database \cite{SCFace} using the EOS \cite{eos} 3DFR algorithm.}
	\label{fig:enlargementexample}
\end{figure}

\textbf{3DFR-based enhancement strategy:} We focus on view-based approaches for enhancing face verification through 3DFR. In particular, we train multiple face recognition systems by considering a gallery enlargement strategy through a single 3DFR method for each neural network to make the system robust to pose variations (Algorithm \ref{alg:projection}). Specifically, we use it on each 3D template generated from training mugshots to project the face in multiple view angles (Figure \ref{fig:enlargementexample}). Then, we train each neural network with all view representations to aid the task of learning how to extract useful information for face verification from non-frontal poses. Concerning the evaluation of each face recognition system at the inference stage, we only compare the frontal face of each subject with the corresponding set of probe images.
The computational cost introduced by the gallery enlargement strategy is mainly offline, thus representing a minor issue in many of the application scenarios to which this contribution is intended \cite{la20233d}.

\begin{algorithm}[t]
\caption{Gallery enlargement from a single personalized 3D template.}\label{alg:projection}
\begin{algorithmic}
\scriptsize
\REQUIRE{3D template}
\ENSURE{2D views of the 3D template}
 \STATE $N \gets 30$ \qquad\qquad\quad\enspace\,\hspace{0.2pt}\COMMENT{Maximum absolute azimuth}
 \STATE $M \gets 30$ \qquad\qquad\quad\enspace\hspace{0.3pt}\COMMENT{Maximum absolute elevation}
 \STATE $\textit{offset} \gets 10$ \qquad\qquad\,\COMMENT{Angle offset}
 \STATE $\textit{el} \gets -M$  \qquad\qquad\quad\hspace{0.1pt}\COMMENT{Initial elevation}
 \WHILE{$\textit{el} \leq M$}
  \STATE $\textit{az} \gets -N$ \qquad\quad\enspace\,\hspace{0.2pt}\COMMENT{Initial azimuth} 
  \WHILE{$\textit{az} \leq N$}
   \STATE Project 3D template in current $\textit{az}$ and $\textit{el}$\;
   \STATE $\textit{az} \gets \textit{az}+\textit{offset}$
   \ENDWHILE
   \STATE $\textit{el} \gets \textit{el}+\textit{offset}$
\ENDWHILE
\normalsize
\end{algorithmic}
\end{algorithm} 

\textbf{Face recognition methods:} We consider two state-of-the-art deep learning networks, XceptionNet \cite{chollet2017xception} and VGG19 \cite{simonyan2014very}, with different computational complexities to simulate various application scenarios. In particular, we selected a Siamese architecture \cite{tolosana2018exploring}, which has demonstrated accurate results in face recognition using low-resolution images \cite{lai2021deep}. 

In order to determine the \textit{a posteriori} probability $P(\textrm{match} | X, Y)$ between the representation obtained from the reconstructed 3D reference model $X$ and the probe image $Y$ match, we calculate the similarity between such a pair of images using the Euclidean distance value $d$ between their feature embeddings (\textit{\ie}, the output of the Siamese Network) as follows:

\begin{small}
\begin{center}
\begin{equation}
  P(\textrm{match} | X, Y) = \frac{1}{d+1}
\end{equation}
\end{center}
\end{small}

In particular, we estimate the \textit{a posteriori} probability through the previous formula to limit the range to $]0, 1]$.

\textbf{Fusion methods:} To the best of our knowledge, the score-level fusion between face recognition systems enhanced through various 3DFR algorithms is still missing in the literature. Accordingly, we explore different non-parametric fusion methods \cite{Ross2009} by applying a set of rules to the \textit{a posteriori} probability values, namely the scores predicted by the single Siamese Neural Networks. Hence, the final \textit{a posteriori} probability obtained from the comparison between a reference and a probe represents a combination of such scores.

Here, we introduce a common notation to ease the understanding of such fusion rules. Let us consider the fusion of $N$ classifier scores, where $P_i(\textrm{match} | X, Y)$ represents the score of the $i$-th classifier when comparing $X$ and $Y$. According to this notation, it is possible to compute the fusion scores through the following formulas that represent, in order, the simple average between the scores obtained from the single recognition systems, their maximum, and their minimum:

\begin{small}
\begin{center}
\begin{equation}
avg\_score = \frac{1}{N}\sum\limits_{i = 1}^N {P_i(\textrm{match} | X, Y)} \label{eq}
\end{equation}
\end{center}
\begin{center}
\begin{equation}
max\_score =  max_i(P_i(\textrm{match} | X, Y))
\end{equation}
\end{center}
\begin{center}
\begin{equation}
min\_score =  min_i(P_i(\textrm{match} | X, Y))
\end{equation}
\end{center}
\end{small}


\section{Experimental Framework}
\label{sec:expsettings}

The description of the experimental framework is divided into three parts. 
Section \ref{subsec:database} describes the database used in our experimental framework. Section \ref{subsec:modelset} explains the experimental protocol considered for the configurations regarding the face recognition systems. It is important to highlight that, as indicated in Section \ref{sec:proposed}, no additional data was used to re-train the 3DFR modules' parameters, nor the one described in Section \ref{subsec:database}. We consider the original versions of the 3DFR algorithms available in the corresponding GitHub repositories. Finally, Section \ref{subsec:performance} discusses the analyzed performance metrics.

\subsection{Database}
\label{subsec:database}

State-of-the-art face recognition systems, even the ones in the 2D domain, achieve nearly perfect performance on traditional benchmark databases \cite{fuad2021recent,melzi2024frcsyn}. However, the most significant advantage of 3DFR algorithms in recognition tasks is in more challenging scenarios, such as surveillance \cite{la20223d}.

Hence, we use the SCface database \cite{SCFace}, containing both high-quality mugshot-like images and lower-quality RGB and grayscale surveillance images of 130 subjects. The surveillance images were acquired through five different camera models at three varying distances from the subject, ranging from 1 to 4.2 meters. Each subject was captured once for every possible camera-distance combination. The observed head poses are typically found in surveillance footage, with the camera slightly above the subject's head \cite{metadataSurvey}. Therefore, SCface is considered to analyze the effect of different quality and resolution cameras on face recognition performance and the robustness to different distances \cite{la20233d}, thus making this database suitable for evaluating interoperability across settings using data from specific cameras or at certain distances. We refer to these evaluations as "cross-settings" experiments, while experiments involving training and testing data from the same camera and distance are referred to as "intra-settings".

Concretely, for both experimental protocols, we use RGB samples related to 25 identities (\textit{\ie}, about 20\% of the subjects) as the test set, while the samples associated with the remaining 105 subjects are further divided, using 90\% as training samples and 10\% as validation ones.
In any experiment concerning the single camera-distance settings for training and test sets, we perform such divisions randomly to limit the possible bias on performance and introduce a face detection stage following the setting used in the same database in \cite{yang2017discriminative}.

\subsection{Face Recognition Systems: Setup}
\label{subsec:modelset}



\begin{figure}[t]
\centering
\includegraphics[width=0.66\linewidth]{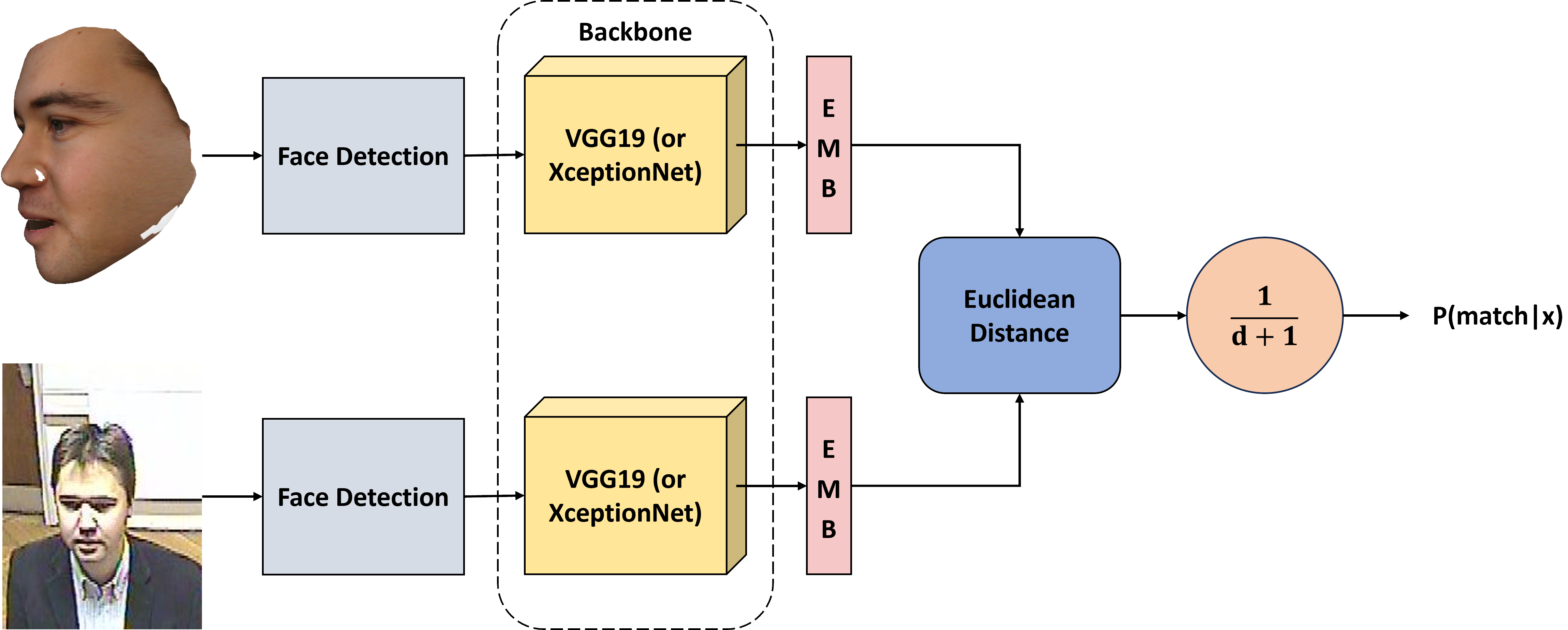}
\caption{Example of the proposed face verification system using a Siamese architecture, introducing as input the probe image and the facial representation in a non-frontal view obtained from a mugshot image through the EOS method \cite{eos}. The images are related to a subject in the SCface database \cite{SCFace}. EMB refers to the feature embeddings obtained from the backbone (\textit{\ie}, VGG19 or XceptionNet). Only frontal views are used in the final inference stage.}\label{fig:subsystem_verification}
\end{figure}

All Siamese networks are pre-trained on the LFW database \cite{LFW}. Then, a fine-tuning through the training set of the SCFace is done, using a validation set for early stopping, according to partitions described in Section \ref{subsec:database}. As shown in Figure \ref{fig:subsystem_verification}, the inputs to the networks are the probe image and the representations obtained from a 3D template of the face related to the claimed identity. For comparability reasons, all the models have been trained on the images resampled at a resolution of $128 \times 128$, on up to 256 epochs with the patience of 5 epochs, batches of 128 triplets, and the Adam optimizer with a learning rate equal to 0.001. Note that this configuration does not necessarily represent the best parameters and preprocessing stages; rather, it is a general configuration that might be further improved in future work.

\subsection{Performance Evaluation} \label{subsec:performance}

After training the individual face recognition systems using information from the different 3DFR algorithms, we investigate the potential of combining them through a correlation analysis between the sets of scores obtained using the test data in intra-setting experiments. In particular, we evaluate the linear correlation between pairs of face recognition systems composed of the same network architecture but enhanced by two different 3DFR algorithms. Therefore, we discuss the complementary information provided by the different 3DFR algorithms in relation to the obtained Pearson Correlation Coefficient (PCC).

Then, we pursue two sets of experiments to evaluate the effectiveness of the fusion methods (\textit{\ie}, the fusion of face recognition systems trained with information provided by different 3DFR algorithms) in the RGB domain, namely, intra-setting and cross-setting. 
The first one consists of training and testing the single systems on data acquired by a specific camera and at a fixed distance from the subjects, therefore evaluating how models behave under controlled conditions. The cross-setting protocol involves training and testing the systems on images acquired with different cameras and distances.

In particular, we evaluate the reliability of the examined fusion methods through metrics commonly used in face recognition. 
From the matching scores, we compute the percentage of False Match Rate (FMR), \textit{\ie}, the rate of non-matching pairs of samples classified as match, and the percentage of False Non-Match Rate (FNMR), \textit{\ie}, the rate of matching pairs of samples that have been incorrectly classified, at various threshold values. This allows us to obtain the Receiver Operating Characteristic (ROC) curve as a function of FMR and $1-\textrm{FNMR}$ and, therefore, the Area Under the Curve (AUC). 
Similarly, from the distributions of the scores related to matching and non-matching identities, we also assess the effect size through Cohen's $d$ to further highlight the discriminatory ability of the evaluated systems. In fact, even if two verification systems report similar AUC values, Cohen's $d$ can also provide information about the magnitude of the differences between the score distributions.
In addition to these performance metrics, we also include for completeness the Equal Error Rate (EER), commonly used for assessing the performance of a biometric system as the percentage of mistakes when the proportions of errors on matching identities and non-matching ones are balanced. Finally, we include the $\%\textrm{FNMR}@\textrm{FMR}=1\%$ ($\%\textrm{FNMR}$ at FMR equal to $1\%$) and $\%\textrm{FMR}@\textrm{FNMR}=1\%$ ($\%\textrm{FMR}$ at FNMR equal to $1\%$) to assess the performance under stringent accuracy constraints, such as high-security and high-usable applications, respectively. 

\section{Results} \label{sec:results}
This section reports the results obtained through the previously described experimental setup. 
Section \ref{subsec:corr} provides an analysis of the correlation between the single classification models enhanced through different 3DFR methods. Section \ref{subsec:intra} describes the outcome of the intra-setting analysis. Finally, Section \ref{subsec:cross} shows the performance obtained through the cross-setting analysis. 



\subsection{Correlation Analysis}\label{subsec:corr}

\begin{figure}[t]
\centering
\includegraphics[width=0.63\linewidth]{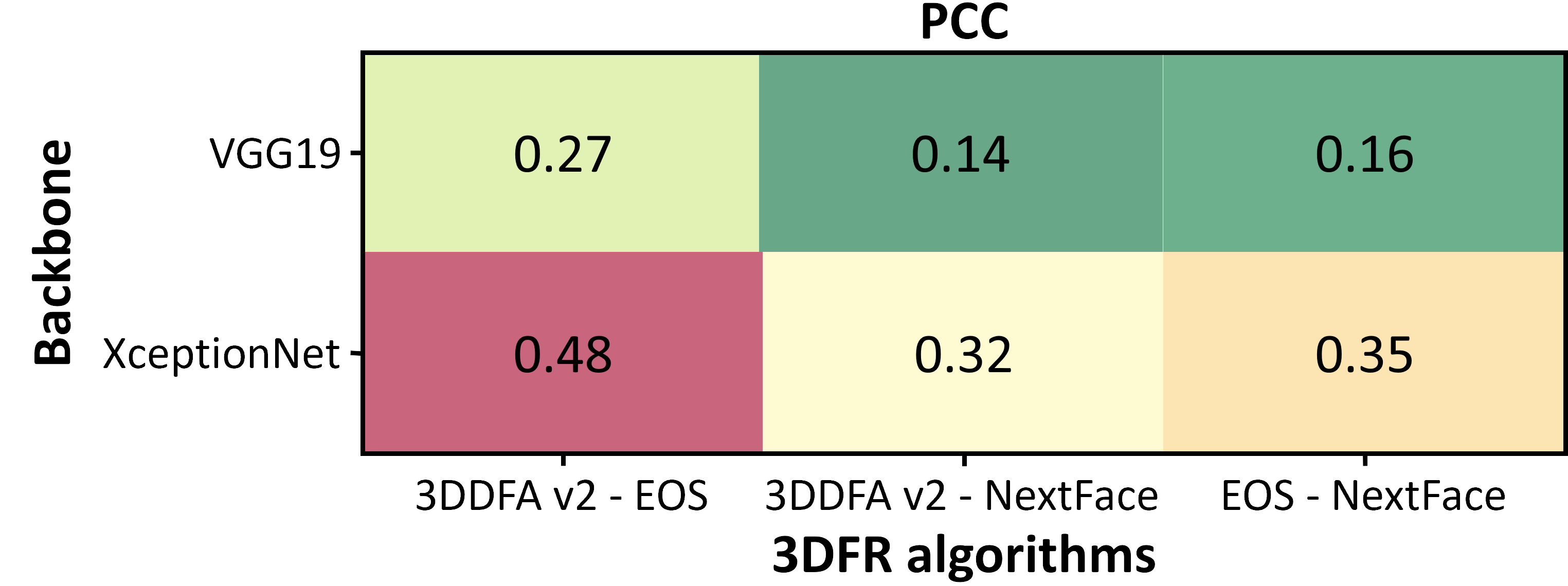}
\caption{Pearson correlation coefficient (PCC) between the set of scores obtained in the intra-setting experiments from pairs of face recognition systems based on the same backbone but enhanced by different 3DFR algorithms.}\label{fig:corr}
\end{figure}

Figure \ref{fig:corr} shows that recognition systems based on the VGG19 backbone are mainly poorly correlated, with the highest correlation obtained from the systems enhanced through 3DDFA v2 and EOS (0.27). Despite the higher values observed, a similar trend is visible in systems based on XceptionNet. This was expected due to the highest quality of the 3D templates generated by 3DDFA v2 and EOS algorithms, as can be seen in Figure \ref{fig:3Dmodels}. 
Hence, the general weak correlation highlighted by this analysis enforces the hypothesis of potential complementary information provided by different 3DFR algorithms. However, despite providing clues on the linear correlation between the scores of pairs of systems, the PCC needs to be paired with an analysis of the effectiveness of the individual models to observe if a low correlation could be related to a relevant difference in the overall performance or mostly to differences in single types of errors, which could therefore be exploited through a combination between them. 


\subsection{Intra-Setting Analysis}\label{subsec:intra}

Table \ref{tab:intraRGB} reports the average performance values in the intra-setting scenario (\textit{\ie}, fixed camera-distance settings) for the individual 3DFR algorithms (3DDFA v2, EOS, and NextFace) and their fusion (Avg, Min, and Max). The best values for each metric and backbone (VGG19 and XceptionNet) are highlighted in bold within the single 3DFR-enhanced models and the fusion approach.

\begin{table*}[t]
\centering
\caption{Average results in the intra-settings context. For each metric and backbone (VGG19 and XceptionNet), we highlight the best result achieved in the individual system (3DDFA v2, EOS, or NextFace) and the best fusion approach (Avg, Min, or Max).}
\resizebox{0.9\textwidth}{!}{%
\begin{tabular}{c|ccccc|ccccc|}
\cline{2-11}
 &
  \multicolumn{5}{c|}{\textbf{VGG19 \cite{simonyan2014very} }} &
  \multicolumn{5}{c|}{\textbf{XceptionNet \cite{chollet2017xception}}}  \\ \hline
\multicolumn{1}{|c|}{\multirow{2}{*}{\textbf{Method}}} &
  \textbf{AUC} &
  \textbf{EER} &
  \textbf{Cohen's} &
  \textbf{\%FMR at} &
  \textbf{\%FNMR at} &
  \textbf{AUC} &
  \textbf{EER} &
  \textbf{Cohen's} &
  \textbf{\%FMR at} &
  \textbf{\%FNMR at}  \\
\multicolumn{1}{|c|}{} &
   \textbf{{[}\%{]}} &
   \textbf{{[}\%{]}} &
  \textbf{d} &
  \textbf{FNMR=1\%} &
  \textbf{FMR=1\%} &
   \textbf{{[}\%{]}} &
   \textbf{{[}\%{]}} &
  \textbf{d} &
  \textbf{FNMR=1\%} &
  \textbf{FMR=1\%}  \\ \hline
\multicolumn{1}{|c|}{\textbf{Baseline}} &
  77.27 &
  15.52 &
  0.91 &
  85.37 &
  87.42 &
  71.89 &
  22.77 &
  0.79 &
  81.37 &
  94.57  \\ \hline
\multicolumn{1}{|c|}{\textbf{3DDFA v2 \cite{guo2020towards}}} &
  81.51 &
  \textbf{12.25} &
  1.09 &
  \textbf{75.72} &
  \textbf{88.75} &
  74.99 &
  \textbf{16.33} &
  0.77 &
  86.83 &
  91.85  \\
\multicolumn{1}{|c|}{\textbf{EOS \cite{eos}}} &
  \textbf{82.06} &
  12.50 &
  \textbf{1.21} &
  80.46 &
  90.66 &
  \textbf{78.18} &
  \textbf{16.33} &
  \textbf{0.92} &
  \textbf{86.43} &
  \textbf{90.68}  \\
\multicolumn{1}{|c|}{\textbf{NextFace \cite{dib2021practical}}} &
  72.58 &
  17.67 &
  0.78 &
  86.45 &
  93.99 &
  71.02 &
  25.83 &
  0.62 &
  90.09 &
  95.58 \\ \hline
\multicolumn{1}{|c|}{\textbf{Fusion (Avg)}} &
  \textbf{86.94} &
  \textbf{10.78} &
  \textbf{1.53} &
  \textbf{56.25} &
  \textbf{80.41} &
  \textbf{79.92} &
  \textbf{15.67} &
  \textbf{1.04} &
  71.57 &
  \textbf{90.19}  \\
\multicolumn{1}{|c|}{\textbf{Fusion (Min)}} &
  84.37 &
  13.36 &
  1.02 &
  84.49 &
  81.78 &
  76.52 &
  18.17 &
  0.61 &
  95.05 &
  91.27  \\
\multicolumn{1}{|c|}{\textbf{Fusion (Max)}} &
  74.07 &
  11.39 &
  1.08 &
  68.12 &
  93.03 &
  77.95 &
  17.75 &
  0.99 &
  \textbf{71.26} &
  94.99  \\ \hline
\end{tabular}%
}
\label{tab:intraRGB}
\end{table*}


We can observe that, in terms of AUC and Cohen's $d$, the best-performing single models are those enhanced through the EOS algorithm. For example, for the VGG19 backbone, the performance achieved when including information from the EOS algorithm is $82.06\%$ AUC, an absolute improvement of $4.79\%$ AUC in comparison with the Baseline system ($77.27\%$ AUC), proving to be 3DFR algorithms a good solution to increase the performance in video surveillance scenarios. A similar trend is observed for XceptionNet. However, for some specific operational points, for example, EER, $\%\textrm{FMR}@\textrm{FNMR}=1\%$, and $\%\textrm{FNMR}@\textrm{FMR}=1\%$, and the VGG19 backbone, the information provided by the 3DDFA v2 algorithm performs better than the EOS algorithm, enforcing the hypothesis that the fusion of face verification systems trained with different 3DFR algorithms may be beneficial.

\begin{figure}[t]
	\centering
	\includegraphics[width=0.64\linewidth]{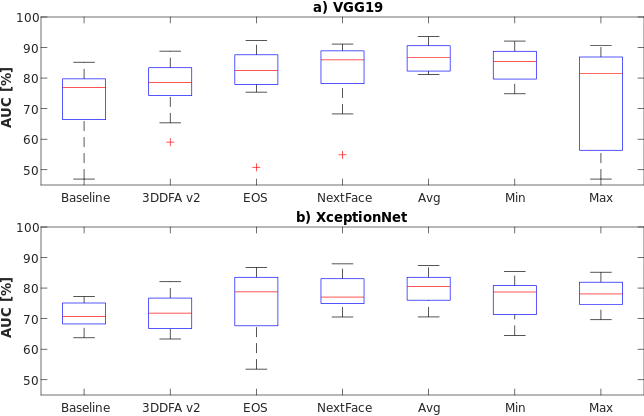}
	\caption{AUC values obtained for the intra-setting scenario for the different camera-distance configurations using VGG19 \cite{simonyan2014very} (a) and XceptionNet \cite{chollet2017xception} (b) as backbones.}
	\label{fig:aucs_intra_rgb}
\end{figure}

Analyzing the proposed fusion (Avg, Min, and Max) of different 3DFR algorithms, we can observe that, in general, the Avg score-level fusion achieves the best results, revealing improved performance both at a global (\textit{\ie}, AUC and Cohen's $d$) and local level (\textit{\ie}, EER, $\%\textrm{FMR}@\textrm{FNMR}=1\%$, and $\%\textrm{FNMR}@\textrm{FMR}=1\%$). For example, for the VGG19 backbone, the Avg score-level fusion achieves an AUC of $86.94\%$, an absolute improvement of $9.67\%$ and $4.88\%$ AUC compared to the Baseline system and the best individual 3DFR algorithm, respectively. 
In particular, this fusion rule shows improvements in the separation between the score distributions related to genuine and impostor subjects, also solving the issues related to the worst performance at challenging thresholds after introducing the 3DFR algorithm in the pipeline.
Furthermore, it is worth noting that this fusion rule is also able to deal with the variability in performance observed by the 3DFR-enhanced recognition systems between the different camera-distance settings. This can be seen in Figure \ref{fig:aucs_intra_rgb}. 
The more stable results across the experiments represent an important feature for the aimed application field \cite{la20233d}, and this is especially achieved with the proposed Avg score-level fusion.

Furthermore, as can be seen in Table \ref{tab:intraRGB} and Figure \ref{fig:aucs_intra_rgb}, the performance of the deep neural networks trained with mugshots or enhanced with single 3DFR algorithms is more robust when employing the VGG19 architecture as the backbone. 
This outcome was expected, considering the significantly higher performance of this network in face recognition tasks \cite{gwyn2021face,kumar2021masked}.
Moreover, the fusion methods related to such architecture also reveal the greatest improvement with respect to the single systems.
Regarding the different fusion methods studied, the minimum and the maximum rules are generally less performing, as they inherently tend to produce more false matches and false non-matches, respectively.

\begin{figure}[t]
	\centering
	\includegraphics[width=0.49\linewidth]{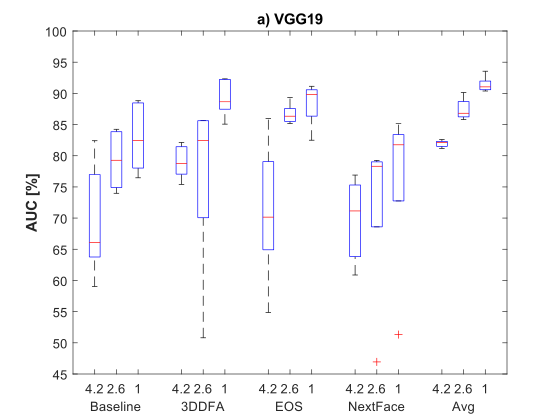}
	\includegraphics[width=0.49\linewidth]{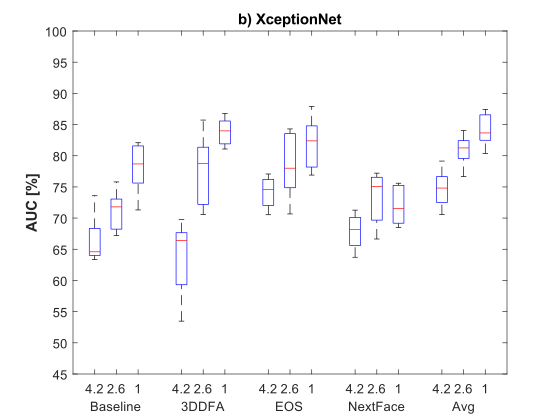}
        \caption{Summary of \textit{AUC} values obtained from all the intra-setting camera-distance configurations (\textit{\ie}, 4.2 meters, 2.6 meters, and 1 meter) through the system based on the VGG19 \cite{simonyan2014very} (a) and XceptionNet \cite{chollet2017xception}  (b), trained with mugshots, single 3DFR algorithms, and fusion of them through the Average rule (\textit{Avg}).}
	\label{fig:aucs_intra_distance}
\end{figure}

Finally, for completeness, we analyze in Figure \ref{fig:aucs_intra_distance} the robustness of the proposed systems with an increasing distance of the camera during the acquisition (\textit{\ie}, 4.2 meters, 2.6 meters, and 1 meter). As expected, both the systems trained through single 3DFR algorithms and their fusion through the Avg rule suffer with an increasing acquisition distance. Similar to our previous analysis, we can observe that the single 3DFR algorithms show differences in enhancement capability, revealing that there is not a single best choice for all settings and, therefore, further highlighting a certain degree of fusion between 3D templates, as can be seen with the Avg score-level fusion.

\subsection{Cross-Setting Analysis}\label{subsec:cross}

Table \ref{tab:crossAllRGB} shows the results achieved when considering different acquisition distances (cross-distance setting) and surveillance cameras (cross-camera setting) between training and test sets.
As expected, both single 3DFR-enhanced and fused systems suffer from a significant decay in performance compared to the intra-setting scenario, remarking how challenging the task is. For example, for the VGG19 backbone and Avg score-level fusion, the AUC decreases from $86.94\%$ in the intra-setting scenario to $74.55\%$ in the cross-setting scenario. Still, the Avg fusion rule is confirmed to be able to improve the overall performance, outperforming the single systems and the Baseline one from both a global and a local perspective. 
A notable finding is that, unlike the intra-setting scenario, the XceptionNet backbone enhanced with the considered 3DFR algorithms achieves better generalization capability in terms of AUC and EER than the VGG19 one.

\begin{table*}[t]
\centering
\caption{Average results in cross-settings context. For each metric and backbone (VGG19 and XceptionNet), we highlight the best result achieved in the individual system (3DDFA v2, EOS, NextFace) and the best fusion approach (Avg, Min, or Max).}
\resizebox{0.9\textwidth}{!}{%
\begin{tabular}{c|ccccc|ccccc|}
\cline{2-11}
 &
  \multicolumn{5}{c|}{\textbf{VGG19 \cite{simonyan2014very} }} &
  \multicolumn{5}{c|}{\textbf{XceptionNet \cite{chollet2017xception} }} \\ \hline
\multicolumn{1}{|c|}{\multirow{2}{*}{\textbf{Method}}} &
  \textbf{AUC} &
  \textbf{EER} &
  \textbf{Cohen's} &
  \textbf{\%FMR at} &
  \textbf{\%FNMR at} &
  \textbf{AUC} &
  \textbf{EER} &
  \textbf{Cohen's} &
  \textbf{\%FMR at} &
  \textbf{\%FNMR at} \\
\multicolumn{1}{|c|}{} &
  \textbf{{[}\%{]}} &
  \textbf{{[}\%{]}} &
  \textbf{d} &
  \textbf{FNMR=1\%} &
  \textbf{FMR=1\%} &
  \textbf{{[}\%{]}} &
  \textbf{{[}\%{]}} &
  \textbf{d} &
  \textbf{FNMR=1\%} &
  \textbf{FMR=1\%} \\ \hline
  \multicolumn{1}{|c|}{\textbf{Baseline}} &
  67.30 &
  36.27 &
  0.55 &
  99.24 &
  98.15 &
  68.97 &
  36.21 &
  0.63 &
  91.81 &
  96.53  \\ \hline
\multicolumn{1}{|c|}{\textbf{3DDFA v2 \cite{guo2020towards}}} &
  68.84 &
  20.69 &
  \textbf{0.64} &
  95.23 &
  94.57 &
  70.95 &
  19.70 &
  \textbf{0.69} &
  \textbf{95.22} &
  92.53 \\
\multicolumn{1}{|c|}{\textbf{EOS \cite{eos}}} &
  \textbf{70.13} &
  \textbf{19.70} &
  0.63 &
  \textbf{93.16} &
  \textbf{93.87} &
  \textbf{71.21} &
  \textbf{19.09} &
  0.61 &
  96.57 &
  \textbf{91.58} \\
\multicolumn{1}{|c|}{\textbf{NextFace \cite{dib2021practical}}} &
  64.91 &
  22.29 &
  0.47 &
  93.71 &
  96.90 &
  68.47 &
  23.46 &
  0.57 &
  95.99 &
  94.97 \\ \hline
\multicolumn{1}{|c|}{\textbf{Fusion (Avg)}} &
  \textbf{74.55} &
  \textbf{19.18} &
  \textbf{0.86} &
  \textbf{84.70} &
  \textbf{90.74} &
  \textbf{75.77} &
  16.44 &
  \textbf{0.86} &
  \textbf{83.41} &
  \textbf{89.84} \\
\multicolumn{1}{|c|}{\textbf{Fusion (Min)}} &
  70.49 &
  20.14 &
  0.58 &
  96.38 &
  91.50 &
  70.70 &
  17.24 &
  0.57 &
  96.67 &
  91.18 \\
\multicolumn{1}{|c|}{\textbf{Fusion (Max)}} &
  68.04 &
  20.57 &
  0.66 &
  88.18 &
  95.66 &
  74.82 &
  \textbf{14.96} &
  \textbf{0.86} &
  84.13 &
  92.89 \\ \hline
\end{tabular}}
\label{tab:crossAllRGB}
\end{table*}
To summarize, despite the expected performance degradation observed in the challenging cross-setting experiments, the proposed fusion of face verification systems trained with complementary information from different 3DFR algorithms demonstrates significant improvement compared to the Baseline and individual 3DFR-enhanced systems. Therefore, the proposed fusion strategy represents a good solution for video surveillance contexts.
The effects of the differences between the technical characteristics of the surveillance cameras and the impact of the different acquisition distances should be further investigated in future research to highlight their impact on system reliability and, therefore, the suitability in specific application scenarios.

\section{Conclusions}\label{sec:conclusion}

In this paper, we investigate the complementary information provided by three state-of-the-art 3D face reconstruction (3DFR) algorithms and the effectiveness of score-level fusion in leveraging this information to improve face recognition in a video surveillance scenario.
We first assess this complementarity by analyzing the linear correlation between the scores produced by different face verification systems, each utilizing the same Siamese Neural Network, using one between the two examined backbones, but enhanced by a different 3DFR algorithm. 
Then, we also employ three non-parametric fusion methods to combine the individual systems based on the same backbone.

Therefore, we evaluate the robustness of the proposed approach using a common experimental setup when dealing with data acquired in intra- and cross-setting, involving variations in acquisition distance and surveillance camera. This allows us to explore different application contexts: the intra-setting experiments simulate a context in which the designer knows the data characteristics and has access to representative data for training the recognition system; the cross-setting experiments simulate a more unfavorable application context in which data characteristics are unknown or representative data cannot be obtained.

The results obtained from our experiments confirm the initial hypothesis that deep-learning face recognition systems can extract distinct information from different state-of-the-art 3DFR algorithms, and the score-level fusion methods effectively leverage this information to enhance inference robustness. 
This improvement can be observed even on probe images obtained from surveillance cameras and at distances different from the ones related to training data.
Therefore, this approach could represent a valid tool for improving facial recognition in challenging surveillance scenarios like the one considered in this work.

Future studies should explore other fusion approaches, such as parametric rules and those based on machine-learning models, which may better exploit the observed complementarity. 
Similarly, more 3DFR algorithms, enhancement approaches, and deep neural networks could be included in the study, and the impact of the proposed approach on different pre-processing settings should be investigated. Finally, the effects of the differences in terms of acquisition distance and surveillance camera between training and test data should be further investigated.

This preliminary study provides valuable insights into the effectiveness of fusion methods in exploiting the complementarity among 3D face reconstruction algorithms within video surveillance contexts. We believe this initial exploration holds promise and could pave the way for developing robust multi-modal face recognition systems capable of recognizing never-seen-before subjects, thereby aiding forensic practitioners and security personnel in identifying criminals and finding missing persons.

\section*{Acknowledgment}
Funding from Cátedra ENIA UAM-VERIDAS en IA Responsable (NextGenerationEU PRTR TSI-100927-2023-2) and project BBforTAI (PID2021-127641OB-I00 MICINN/FEDER). The work has been conducted within the sAIfer Lab and the ELLIS Unit Madrid.


%
%
\bibliographystyle{splncs04}
\bibliography{main}

\begin{thebibliography}{10}
\providecommand{\url}[1]{\texttt{#1}}
\providecommand{\urlprefix}{URL }
\providecommand{\doi}[1]{https://doi.org/#1}

\bibitem{2D3Dsurvey}
Abate, A.F., Nappi, M., Riccio, D., Sabatino, G.: {2D} and {3D} face recognition: A survey. Pattern Recognition Letters  \textbf{28}(14),  1885--1906 (2007), image: Information and Control

\bibitem{ICPrecognition}
Amor, B.B., Ouji, K., Ardabilian, M., Chen, L.: {3D} face recognition by icp-based shape matching. LIRIS Lab, Lyon Research Center for Images and Intelligent Information Systems, UMR  \textbf{5205} (2005)

\bibitem{10.1145/3596711.3596730}
Blanz, V., Vetter, T.: A Morphable Model For The Synthesis Of {3D} Faces. Association for Computing Machinery, New York, NY, USA, 1 edn. (2023)

\bibitem{Cao2020}
Cao, J., Hu, Y., Zhang, H., He, R., Sun, Z.: Towards high fidelity face frontalization in the wild. International Journal of Computer Vision  \textbf{128}(5),  1485--1504 (2020)

\bibitem{metadataSurvey}
Castro, H.F., Cardoso, J.S., Andrade, M.T.: A systematic survey of ml datasets for prime cv research areas—media and metadata. Data  \textbf{6}(2), ~12 (2021)

\bibitem{chollet2017xception}
Chollet, F.: Xception: Deep learning with depthwise separable convolutions. In: Proceedings of the IEEE Conference on Computer Vision and Pattern Recognition. pp. 1251--1258 (2017)

\bibitem{concas2022experimental}
Concas, S., Gao, J., Cuccu, C., Orr{\`u}, G., Feng, X., Marcialis, G.L., Puglisi, G., Roli, F.: Experimental results on multi-modal deepfake detection. In: International Conference on Image Analysis and Processing. pp. 164--175. Springer (2022)

\bibitem{robust}
Creusot, C., Pears, N., Austin, J.: A machine-learning approach to keypoint detection and landmarking on {3D} meshes. International Journal of Computer Vision  \textbf{102}(1),  146--179 (2013)

\bibitem{dass2005principled}
Dass, S.C., Nandakumar, K., Jain, A.K.: A principled approach to score level fusion in multimodal biometric systems. In: International Conference on Audio- and Video-based Biometric Person Authentication. pp. 1049--1058. Springer (2005)

\bibitem{deandres2024frcsyn}
DeAndres-Tame, I., Tolosana, R., Melzi, P., Vera-Rodriguez, R., Kim, M., Rathgeb, C., Liu, X., Morales, A., Fierrez, J., Ortega-Garcia, J., et~al.: {Second Edition FRCSyn challenge at CVPR 2024:} face recognition challenge in the era of synthetic data. In: Proc. IEEE/CVF Conference on Computer Vision and Pattern Recognition (2024)

\bibitem{dib2021practical}
Dib, A., Bharaj, G., Ahn, J., Th{\'e}bault, C., Gosselin, P., Romeo, M., Chevallier, L.: Practical face reconstruction via differentiable ray tracing. In: Computer Graphics Forum. vol.~40, pp. 153--164. Wiley Online Library (2021)

\bibitem{fuad2021recent}
Fuad, M.T.H., Fime, A.A., Sikder, D., Iftee, M.A.R., Rabbi, J., Al-Rakhami, M.S., Gumaei, A., Sen, O., Fuad, M., Islam, M.N.: Recent advances in deep learning techniques for face recognition. IEEE Access  \textbf{9},  99112--99142 (2021)

\bibitem{geng2020towards}
Geng, Z., Cao, C., Tulyakov, S.: Towards photo-realistic facial expression manipulation. International Journal of Computer Vision  \textbf{128},  2744--2761 (2020)

\bibitem{SCFace}
Grgic, M., Delac, K., Grgic, S.: Scface--surveillance cameras face database. Multimedia Tools and Applications  \textbf{51}(3),  863--879 (2011)

\bibitem{guo2020towards}
Guo, J., Zhu, X., Yang, Y., Yang, F., Lei, Z., Li, S.Z.: Towards fast, accurate and stable {3D} dense face alignment. In: Proceedings of the European Conference on Computer Vision (ECCV) (2020)

\bibitem{gwyn2021face}
Gwyn, T., Roy, K., Atay, M.: Face recognition using popular deep net architectures: A brief comparative study. Future Internet  \textbf{13}(7), ~164 (2021)

\bibitem{HanJain2012}
Han, H., Jain, A.K.: {3D} face texture modeling from uncalibrated frontal and profile images. In: 2012 IEEE Fifth International Conference on Biometrics: Theory, Applications and Systems (BTAS). pp. 223--230. IEEE (2012)

\bibitem{hassner2015effective}
Hassner, T., Harel, S., Paz, E., Enbar, R.: Effective face frontalization in unconstrained images. In: Proceedings of the IEEE Conference on Computer Vision and Pttern Recognition. pp. 4295--4304 (2015)

\bibitem{3Dstereo}
Heike, C.L., Upson, K., Stuhaug, E., Weinberg, S.M.: {3D} digital stereophotogrammetry: a practical guide to facial image acquisition. Head \& Face Medicine  \textbf{6}(1),  1--11 (2010)

\bibitem{HU201746}
Hu, X., Peng, S., Wang, L., Yang, Z., Li, Z.: Surveillance video face recognition with single sample per person based on {3D} modeling and blurring. Neurocomputing  \textbf{235},  46--58 (2017)

\bibitem{LFW}
Huang, G.B., Mattar, M., Berg, T., Learned-Miller, E.: Labeled faces in the wild: A database for studying face recognition in unconstrained environments. In: Workshop on Faces in 'Real-Life' Images: Detection, Alignment, and Recognition (2008)

\bibitem{eos}
Huber, P., Hu, G., Tena, R., Mortazavian, P., Koppen, P., Christmas, W.J., Ratsch, M., Kittler, J.: A multiresolution {3D} morphable face model and fitting framework. In: Proceedings of the 11th International Joint Conference on Computer Vision, Imaging and Computer Graphics Theory and Applications (2016)

\bibitem{kumar2021masked}
Kumar, M., Mann, R.: Masked face recognition using deep learning model. In: 2021 3rd International Conference on Advances in Computing, Communication Control and Networking (ICAC3N). pp. 428--432. IEEE (2021)

\bibitem{la20233d}
La~Cava, S.M., Orr{\`u}, G., Drahansky, M., Marcialis, G.L., Roli, F.: {3D} face reconstruction: the road to forensics. ACM Computing Surveys (CSUR)  \textbf{56}(3),  1--38 (2023)

\bibitem{la20223d}
La~Cava, S.M., Orr{\`u}, G., Goldmann, T., Drahansky, M., Marcialis, G.L.: {3D} face reconstruction for forensic recognition-a survey. In: 2022 26th International Conference on Pattern Recognition (ICPR). pp. 930--937. IEEE (2022)

\bibitem{lai2021deep}
Lai, S.C., Lam, K.M.: Deep siamese network for low-resolution face recognition. In: 2021 Asia-Pacific Signal and Information Processing Association Annual Summit and Conference (APSIPA ASC). pp. 1444--1449. IEEE (2021)

\bibitem{Liang2018}
Liang, J., Liu, F., Tu, H., Zhao, Q., Jain, A.K.: On mugshot-based arbitrary view face recognition. In: 2018 24th International Conference on Pattern Recognition (ICPR). pp. 3126--3131. IEEE (2018)

\bibitem{marcialis2004fusion}
Marcialis, G.L., Roli, F.: Fusion of appearance-based face recognition algorithms. Pattern Analysis and Applications  \textbf{7},  151--163 (2004)

\bibitem{melzi2024frcsyn}
Melzi, P., Tolosana, R., Vera-Rodriguez, R., Kim, M., Rathgeb, C., Liu, X., DeAndres-Tame, I., Morales, A., Fierrez, J., Ortega-Garcia, J., et~al.: {FRCSyn-onGoing:} benchmarking and comprehensive evaluation of real and synthetic data to improve face recognition systems. Information Fusion  \textbf{107},  102322 (2024)

\bibitem{uncalibrated}
Morales, A., Piella, G., Sukno, F.M.: Survey on {3D} face reconstruction from uncalibrated images. Computer Science Review  \textbf{40},  100400 (2021)

\bibitem{Noore2009}
Noore, A., Singh, R., Vasta, M.: Fusion, Sensor-Level, pp. 616--621. Springer US, Boston, MA (2009)

\bibitem{Osadciw2009}
Osadciw, L., Veeramachaneni, K.: Fusion, Decision-Level, pp. 593--597. Springer US, Boston, MA (2009)

\bibitem{peng2014multimodal}
Peng, J., Abd El-Latif, A.A., Li, Q., Niu, X.: Multimodal biometric authentication based on score level fusion of finger biometrics. Optik  \textbf{125}(23),  6891--6897 (2014)

\bibitem{punyani_evaluation_2017}
Punyani, P., Kumar, A.: Evaluation of fusion at different levels for face recognition. In: 2017 {Int.} {Conference} on {Computing}, {Communication} and {Automation} ({ICCCA}). pp. 1052--1055 (May 2017)

\bibitem{richardson20163d}
Richardson, E., Sela, M., Kimmel, R.: {3D} face reconstruction by learning from synthetic data. In: Proc. Fourth International Conference on {3D} Vision (2016)

\bibitem{Ross2009}
Ross, A., Nandakumar, K.: Fusion, Score-Level, pp. 611--616. Springer US, Boston, MA (2009)

\bibitem{shahreza2024sdfr}
Shahreza, H.O., Ecabert, C., George, A., Unnervik, A., Marcel, S., Di~Domenico, N., Borghi, G., Maltoni, D., Boutros, F., Vogel, J., et~al.: {SDFR:} synthetic data for face recognition competition. In: Proc. IEEE International Conference on Automatic Face and Gesture Recognition (2024)

\bibitem{simonyan2014very}
Simonyan, K., Zisserman, A.: Very deep convolutional networks for large-scale image recognition. arXiv preprint arXiv:1409.1556  (2014)

\bibitem{robustsurvey}
Soltanpour, S., Boufama, B., Wu, Q.J.: A survey of local feature methods for {3D} face recognition. Pattern Recognition  \textbf{72},  391--406 (2017)

\bibitem{sun2022controllable}
Sun, K., Wu, S., Huang, Z., Zhang, N., Wang, Q., Li, H.: Controllable {3D} face synthesis with conditional generative occupancy fields. In: Proc. Advances in Neural Information Processing Systems (2022)

\bibitem{tolosana2018exploring}
Tolosana, R., Vera-Rodriguez, R., Fierrez, J., Ortega-Garcia, J.: Exploring recurrent neural networks for on-line handwritten signature biometrics. IEEE Access  \textbf{6},  5128--5138 (2018)

\bibitem{vishi_evaluation_2018}
Vishi, K., Mavroeidis, V.: An {Evaluation} of {Score} {Level} {Fusion} {Approaches} for {Fingerprint} and {Finger}-vein {Biometrics} (2018)

\bibitem{yang2017discriminative}
Yang, F., Yang, W., Gao, R., Liao, Q.: Discriminative multidimensional scaling for low-resolution face recognition. IEEE Signal Processing Letters  \textbf{25}(3),  388--392 (2017)

\bibitem{Zhang2008}
Zhang, X., Gao, Y., Leung, M.K.: Recognizing rotated faces from frontal and side views: An approach toward effective use of mugshot databases. IEEE Transactions on Information Forensics and Security  \textbf{3}(4),  684--697 (2008)

\bibitem{RotateAndRender}
Zhou, H., Liu, J., Liu, Z., Liu, Y., Wang, X.: Rotate-and-render: Unsupervised photorealistic face rotation from single-view images. In: Proceedings of the IEEE/CVF conference on computer vision and pattern recognition. pp. 5911--5920 (2020)

\end{thebibliography}
\end{document}